\documentclass[10pt,twocolumn,letterpaper]{article}

\usepackage{iccv}
\usepackage{times}
\usepackage{epsfig}
\usepackage{graphicx}
\usepackage{amsmath}
\usepackage{amssymb}

\usepackage{threeparttable}
\usepackage{multirow}
\usepackage{booktabs}
\usepackage{array}
\usepackage{setspace}
\usepackage{url}

\usepackage[accsupp]{axessibility}

\usepackage[pagebackref=false,colorlinks]{hyperref}
\usepackage[all]{hypcap}

\iccvfinalcopy 

\def\iccvPaperID{2755} 

\begin{document}

\title{MFNet: Multi-filter Directive Network for Weakly Supervised \\ Salient Object Detection}

\author{
Yongri Piao\textsuperscript{1\thanks{Equal Contributions}}\qquad Jian Wang\textsuperscript{\rm 1\textsuperscript{$\ast$}}\qquad Miao Zhang\textsuperscript{\rm 1,2\thanks{Corresponding Author}}\qquad  Huchuan Lu\textsuperscript{\rm 1,3} \\
$^{1}$ Dalian University of Technology, China\\
$^{2}$ Key Lab for Ubiquitous Network and Service Software of Liaoning Province,\\
Dalian University of Technology, China\\
$^{3}$ Pengcheng Lab\\
{\tt\small \{yrpiao, miaozhang, lhchuan\}@dlut.edu.cn}\quad
{\tt\small \{dlyimi\}@mail.dlut.edu.cn}
}


\maketitle
\ificcvfinal\thispagestyle{empty}\fi
\pagestyle{empty}  
\thispagestyle{empty} 

\vspace{-3mm}
\begin{abstract}
Weakly supervised salient object detection (WSOD) targets to train a CNNs-based saliency network using only low-cost annotations. Existing WSOD methods take various techniques to pursue single "high-quality" pseudo label from low-cost annotations and then develop their saliency networks. Though these methods have achieved good performance, the generated single label is inevitably affected by adopted refinement algorithms and shows prejudiced characteristics which further influence the saliency networks. In this work, we introduce a new multiple-pseudo- label  framework to integrate more comprehensive and accurate saliency cues from multiple labels, avoiding the aforementioned problem. Specifically, we propose a multi-filter directive network (MFNet) including a saliency network as well as multiple directive filters. The directive filter (DF) is designed to extract and filter more accurate saliency cues from the noisy pseudo labels. The multiple accurate cues from multiple DFs are then simultaneously propagated to the saliency network with a multi-guidance loss. Extensive experiments on five datasets over four metrics demonstrate that our method outperforms all the existing congeneric methods. Moreover, it is also worth noting that our framework is flexible enough to apply to existing methods and improve their performance. 
The code and results of our method are available at \textcolor{magenta}{\url{https://github.com/OIPLab-DUT/MFNet}}.
\end{abstract}

\vspace{-3mm}
\section{Introduction}
With the emergence of convolutional neural networks (CNNs)~\cite{1998Gradient}, a lot of salient object detection (SOD) methods~\cite{2018Detect, 7780449, Wu2019CascadedPD, Zhang2017AmuletAM} based on CNNs have been proposed and broken the records. However, these CNNs-based SOD methods heavily rely on large amounts of hand-labeling data with pixel-level annotations, which are labor-intensive and time-consuming~\cite{Zhang2019MemoryorientedDF}.

Due to the high cost of labeling pixel-level annotations, some promising works have been proposed to explore other low-cost alternatives, including scribble~\cite{Zhang2020WeaklySupervisedSO, Yu2020StructureConsistentWS} and image-level category labels~\cite{Zeng2019MultiSourceWS, Wang2017LearningTD, Li2018WeaklySS}.
Among them, the category label based methods only require category labels for training, and an overwhelming amount of labels for the existence of object categories are already given (\eg ImageNet~\cite{5206848}). Thus, in this paper, we focus on the image-level category label based salient object detection (WSOD\footnote{For convenience, we denote WSOD as methods based on image-level category label in this paper.}).

\begin{figure}
\vspace{-1mm}
\includegraphics[width=1.00\linewidth]{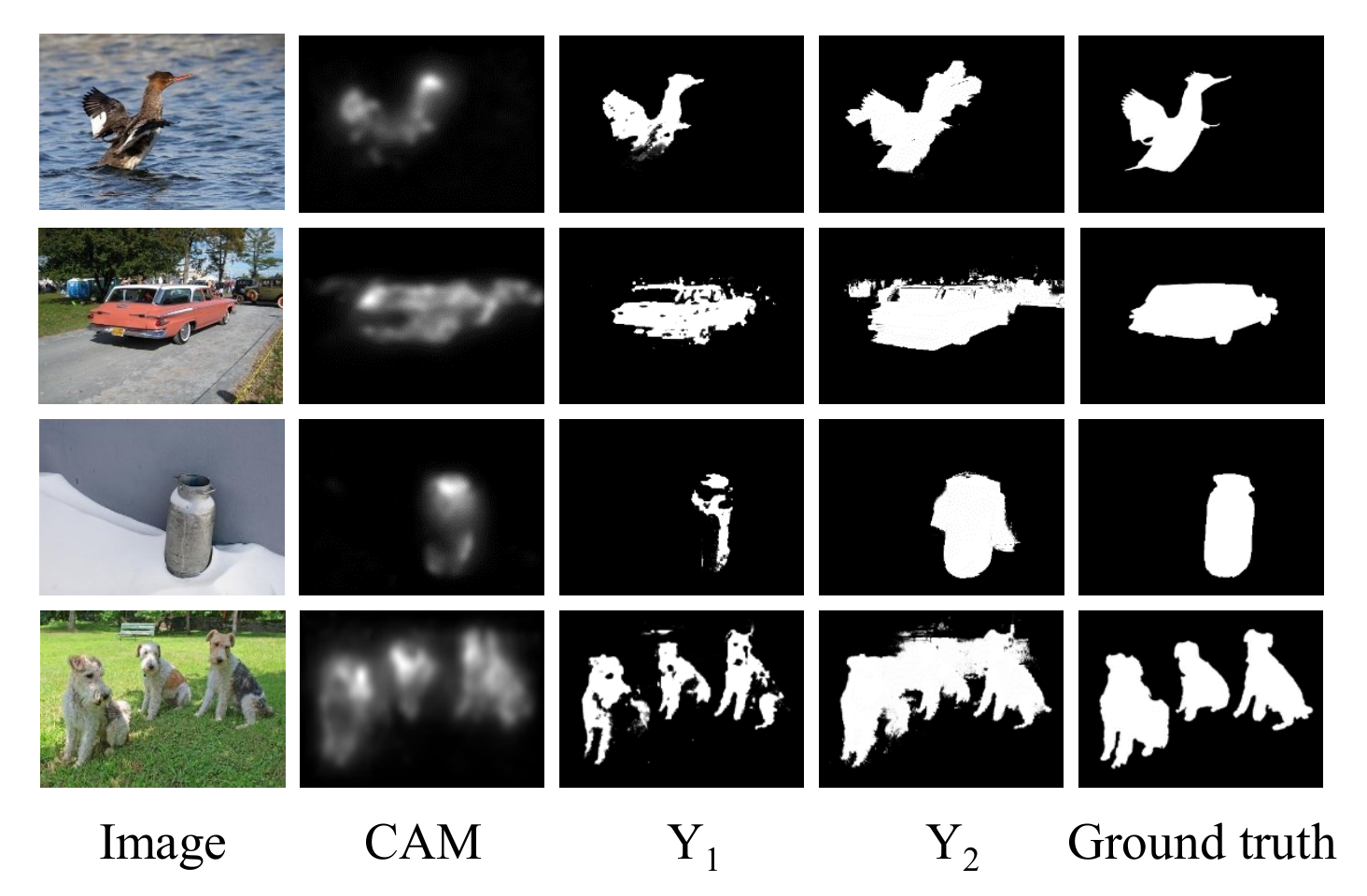}
\vspace{-4mm}
\caption{ Different pseudo labels synthesized by different refinement algorithms on class activation map (CAM), in which $Y_1$ and $Y_2$ represent pseudo labels from pixel-wise~\cite{Araslanov2020SingleStageSS} and superpixel-wise~\cite{Wang2017LearningTD} refinement algorithms, respectively. }
\label{introduction}
\vspace{-2mm}
\end{figure}

Previous works on WSOD proposed various techniques such as global smooth pooling~\cite{Wang2017LearningTD}, multi-source supervisions~\cite{Zeng2019MultiSourceWS} and alternate optimization~\cite{Li2018WeaklySS} to pursue single "high-quality" pseudo label for training their saliency networks. Though these works have achieved good performance, the generated single "high-quality" pseudo label is usually trapped by its prejudiced characteristics due to the different adopted refinement algorithms. For example, the incomplete deficiency ($3^{rd}$ column in Figure {\color{red}\ref{introduction}}) and redundant noise ($4^{th}$ column in Figure {\color{red}\ref{introduction}}).

Instead of pursuing single "high-quality" pseudo labels, we propose to utilize multiple pseudo labels to establish a more robust framework and avoid the negative impacts from the single prejudiced label. To begin with, we adopt two different refinement algorithms, including a pixel-wise one~\cite{Araslanov2020SingleStageSS} and a superpixel-wise one~\cite{Wang2017LearningTD}, to synthesize two different pseudo labels. Both of these two algorithms utilize abundant appearance information in RGB images to perform refinement for class activation maps (CAMs)~\cite{Zhou2016LearningDF}. The pixel-wise one treats each individual pixel as units, takes its class activation score as clues and then infers its neighbor pixels' scores, while the superpixel-wise one takes superpixels as its operation units. As a result, the synthesized pseudo labels $Y_1$ (from pixel-wise algorithm) and $Y_2$ (from superpixel-wise algorithm) describe different characteristics. As is shown in Figure {\color{red}\ref{introduction}}, $Y_1$ provides better detailed information, but is usually trapped in incompleteness, while $Y_2$ can cover more complete objects but introduces more extra noisy information. These observations drive us to explore how to extract and integrate more comprehensive and robust saliency cues from multiple pseudo labels.

The core insight of this work is to adequately excavate the comprehensive saliency cues in multiple pseudo labels and avoid the prejudice of the single label. To be specific, for multiple pseudo labels, we 1) extract abundant accurate multiple saliency cues from multiple noisy labels, and 2) perform integration and propagate the integrated multiple cues to the saliency network. Concretely, our contributions are as follows:

\begin{itemize}
\hyphenpenalty=1000
\tolerance=10
\vspace{-2mm}
\item 	We introduce a new framework to utilize multiple pseudo labels for WSOD, which employs more comprehensive and robust saliency cues in multiple labels to avoid the negative impacts of a single label.
\vspace{-2mm}
\item 	We design a multi-filter directive network (denoted as MFNet), in which multiple directive filters and a multi-guidance loss are proposed to extract and integrate multiple saliency cues from multiple pseudo labels respectively.
\vspace{-2mm}
\item 	Extensive experiments on five benchmark datasets over four metrics demonstrate the superiority of our method as well as the multiple pseudo labels.
\vspace{-2mm}
\item 	We also extend the proposed framework to existing method MSW~\cite{Zeng2019MultiSourceWS} and the prove its effectiveness by achieving 9.1\% improvements over $F_{\beta}^{\omega}$ metric on ECSSD dataset.
\end{itemize}

\vspace{-2mm}
\section{Related Work}
\subsection{Salient Object Detection}
Early researches on salient object detection (SOD) mainly leverage handcrafted features to segment the most salient objects, such as boundary prior~\cite{2013Saliency}, center prior~\cite{Jiang2013SubmodularSR} and so on~\cite{Zhu2014SaliencyOF, 2014Salient}. 
Recently, CNNs-based approaches have yielded a qualitative leap in performance due to the powerful ability of CNNs in extracting informative features. 
Various effective architectures~\cite{7780449, Zhang2017AmuletAM, Wang2017ASR, Luo2017NonlocalDF, Liu2018PiCANetLP} are proposed to enhance the performance of the saliency networks, among them, Liu \etal~\cite{7780449} propose a deep hierarchical saliency network, which can simultaneously learn powerful feature representations, informative saliency cues, and their optimal combination mechanisms from the global view. 
With the development of attention mechanisms, some promising works~\cite{Wu2019CascadedPD, Piao2019DepthInducedMR, Zhang2020SelectSA} are presented to introduce various attention modules to improve the saliency networks, in which Wu \etal~\cite{Wu2019CascadedPD} introduce a cascaded partial decoder framework, utilizing generated relatively precise attention maps to refine high-level features for improving the performance.
In recent years, boundary information is attached much importance, and lots of works~\cite{Liu2019ASP, Su2019SelectivityOI, Zhao2019EGNetEG} propose to explore boundary of the salient objects to predict a more detailed prediction. 
In~\cite{Su2019SelectivityOI}, Su \etal propose an effective Cross Reﬁnement Unit (CRU), which bidirectionally passes messages between the two tasks of salient object detection and edge detection.

Although these methods have achieved promising improvements, a large amount of pixel-level annotations are needed for training their models, which are prohibitively expensive.

\subsection{Weakly Supervised Salient Object Detection}
To achieve a trade-off between labeling efficiency and model performance, weakly supervised salient object detection using low-cost labels is presented.
Wang \etal~\cite{Wang2017LearningTD} first propose to perform salient object detection with image-level category labels and design a foreground inference network (FIN) to infer saliency maps. A global smooth pooling (GSP) is proposed to generate more integrated CAMs from image-level labels, and a new CRF algorithm which provides more accurate refinementis also proposed to giving rise to more effective network training.
In~\cite{Li2018WeaklySS}, Li \etal design a generic alternate optimization framework to progressively refine and update the initial saliency seeds from a traditional SOD method MB+~\cite{Zhang2015MinimumBS}, a conditional random field based graphical model is also introduced to cleanse the noisy pseudo labels.

\begin{figure*}[!t]
\vspace{-1mm}
\includegraphics[width=1.00\linewidth]{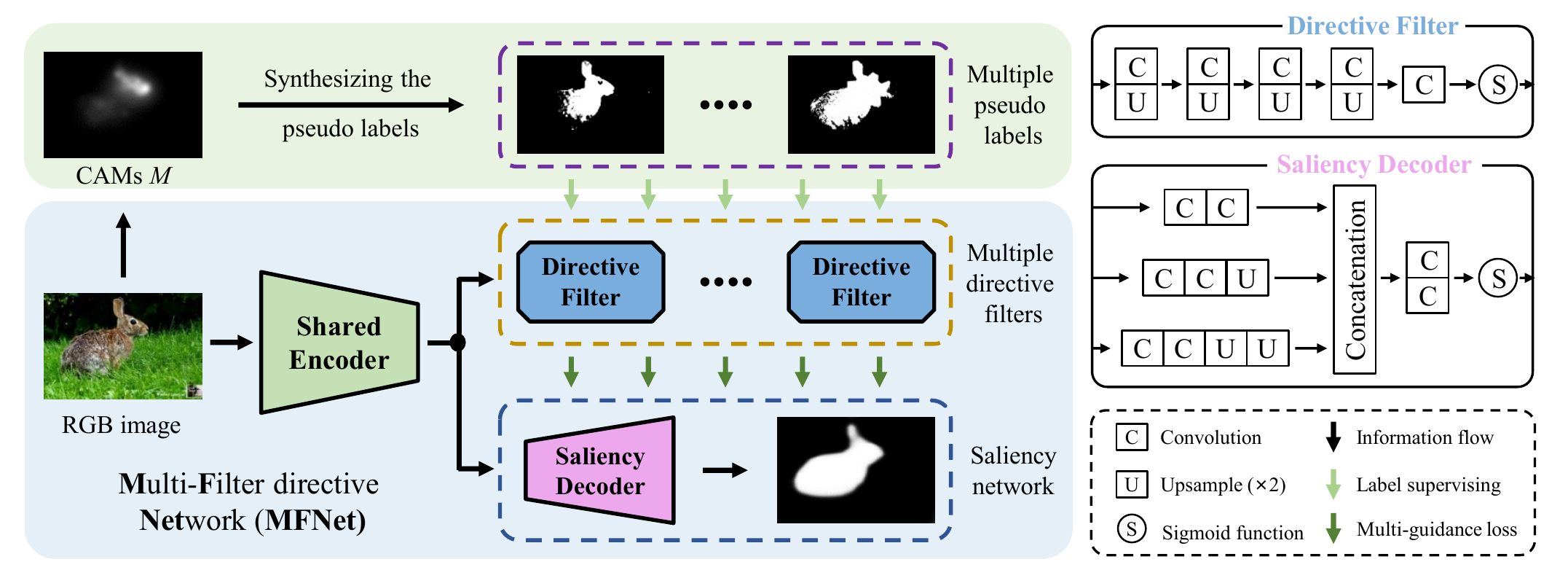}
\setlength{\abovecaptionskip}{0.2cm}
\setlength{\belowcaptionskip}{0.2cm}
\vspace{-4mm}
\caption{ Overall framework of our proposed method. The class activation maps (CAMs)~\cite{Zhou2016LearningDF} are inferred by a trained image classification network, and multiple pseudo labels are synthesized based on it. The proposed MFNet includes two directive filters and a normal encoder-decoder saliency network. The architecture of the saliency decoder and directive filter is illustrated on the right, in which the three inputs of the saliency decoder represent the features from the $3^{rd}$, $4^{th}$ and $5^{th}$ convolution block of the shared encoder. }
\label{framework}
\vspace{-3mm}
\end{figure*}

Different from the previous works, Zeng \etal~\cite{Zeng2019MultiSourceWS} propose that the saliency cues in category labels can be supplemented by caption labels, and design a multi-source weak supervision framework to integrate multiple information in various supervisions. Besides, an attention transfer loss is proposed to transmit supervision signal between networks, and an attention coherence loss is presented to encourage the networks to detect the generally salient regions.
Owe to the abundant saliency information in multi-source weak supervisions, a promising improvement is achieved in ~\cite{Zeng2019MultiSourceWS}.
However, the multi-source framework only integrates the abundant information to generate a single pseudo label, leading that multi-source information cannot be explicitly propagated to the saliency network. \textbf{In conclusion}, the above previous works target to pursue a single "high-quality" pseudo label and then develop saliency networks on it.

Different from the aforementioned works, we hold that the saliency cues in image-level category label can be differently excavated to synthesize multiple pseudo labels. The saliency network developed on these multiple labels can be more robust and avoid the prejudiced effects of single labels.

\vspace{-3mm}
\section{The Proposed Method}
To excavate the comprehensive saliency cues in multiple pseudo labels, we propose a multiple pseudo label framework. As is illustrated in Figure {\color{red}\ref{framework}}, the proposed framework can be divided to two parts:  \textbf{1)} Synthesizing multiple pixel-level pseudo labels on training images given existing image-level classification dataset. \textbf{2)} Developing the proposed multi-filter directive network (\textbf{MFNet}) with the generated multiple labels. In this section, we will introduce the first part in a brief way and then give the detailed descriptions of the second one.

\subsection{Synthesizing Multiple Pseudo Labels}
Based on a image classification network, class activation maps (CAMs)~\cite{Zhou2016LearningDF} build a bridge from image-level category labels to pixel-level pseudo labels, and play a vital role in weakly supervised segmentation tasks. Similar to~\cite{Zeng2019MultiSourceWS, Wang2017LearningTD}, we adopt ImageNet dataset~\cite{5206848} as the training set in this part for the sake of fairness.

For an image classification network, we replace the fully connected layers with a global average pooling (GAP)~\cite{Lin2014NetworkIN} layer and add an extra convolution layer. The GAP layer encourages the classification network to identify the more complete extent of the object. The classification scores $S$ are computed by:

\vspace{-4mm}
\begin{equation}
\begin{split}
S = conv(GAP({F_5})),
\end{split}
\end{equation}
\vspace{-5mm}

\noindent where ${conv(\cdot)}$ represents the new-added convolution layer, and ${F_5}$ represents the features from the last convolution block of the classification network. The classification loss ${L_c}$ in this training stage is as follows:

\vspace{-4mm}
\begin{equation}
\begin{split}
{L_c}(S,{Y_c}) 	& =  - {1 \over C}*\sum\limits_{i = 1}^C  { {y_{ci}}*\log ({{(1 + \exp ( - {s_i}))}^{ - 1}})} \\
							& + (1 - {y_{ci}})*\log ({{\exp ( - {s_i})} \over {1 + \exp ( - {s_i})}}),
\end{split}
\end{equation}
\vspace{-4mm}

\noindent where $C$ indicates the total numbers of category, $y_{ci}$ and ${s_i}$ represent the elements of the category label $Y_c$ and the computed classification scores $S$, respectively.

After the training stage of the classification network is completed, we fix the learned parameters and perform inference on the RGB image of DUTS-Train dataset~\cite{Wang2017LearningTD} to generate class activation maps (CAMs) $M$ as follows:


\vspace{-4mm}
\begin{equation}
\begin{split}
M = \sum\limits_{i = 1}^C {\;norm(relu{{(conv({F_5})}_i}))} *{s_i},
\end{split}
\end{equation}
\vspace{-3mm}

\noindent where ${conv(\cdot)}$ is the aforementioned new-added convolutional layer. $relu(\cdot)$ indicates the relu activation function, and $norm(\cdot)$ represents the normalization function that normalizes the elements in CAMs to [0, 1].

As is mentioned above, we adopt both pixel-wise and superpixel-wise algorithms on CAMs for refinements. The pixel-wise refinement~\cite{Araslanov2020SingleStageSS} takes the class activation score of individual pixel in CAMs as seeds, and infers the scores of its neighbor pixels using the RGB appearance information. On the other hand, superpixel-wise refinement first clusters pixels in a RGB image into superpixels using a clustering algorithm SLIC~\cite{Achanta2012SLICSC} and then performs the similar refinement on superpixels. Same as the previous works~\cite{Zeng2019MultiSourceWS, Wang2017LearningTD, Li2018WeaklySS}, we also adopt CRF~\cite{Krhenbhl2011EfficientI} for a further refinement, which is widely-accepted in the weakly supervised methods.

\subsection{The Multi-filter Directive Network }
As is mentioned above, pseudo labels synthesized from different refinements describe different characteristics, and the saliency networks developed on a single label inevitably suffers from its prejudiced characteristics. Therefore, we target to explore how to effectively leverage the abundant and comprehensive saliency cues in multiple pseudo labels.

A straightforward method to utilize multiple cues is designing a dual decoder architecture as shown in (b) in Figure {\color{red}\ref{ablation_set}}, which introduces two decoders to learn saliency cues from two different pseudo labels respectively. Meanwhile a mutual guidance loss is adopted to integrate multiple saliency cues. We take the averaged predictions of dual decoders as the final saliency prediction in this ease.
However, in this straightforward method, noisy information existing in prejudiced pseudo label may propagate to the saliency network directly, and lead to negative impacts.
To solve the above problems, we propose a multi-filter directive network (MFNet) to effectively integrate the filtered cues from multiple pseudo labels.

\begin{figure}[!t]
\vspace{-0mm}
\includegraphics[width=1.00\linewidth]{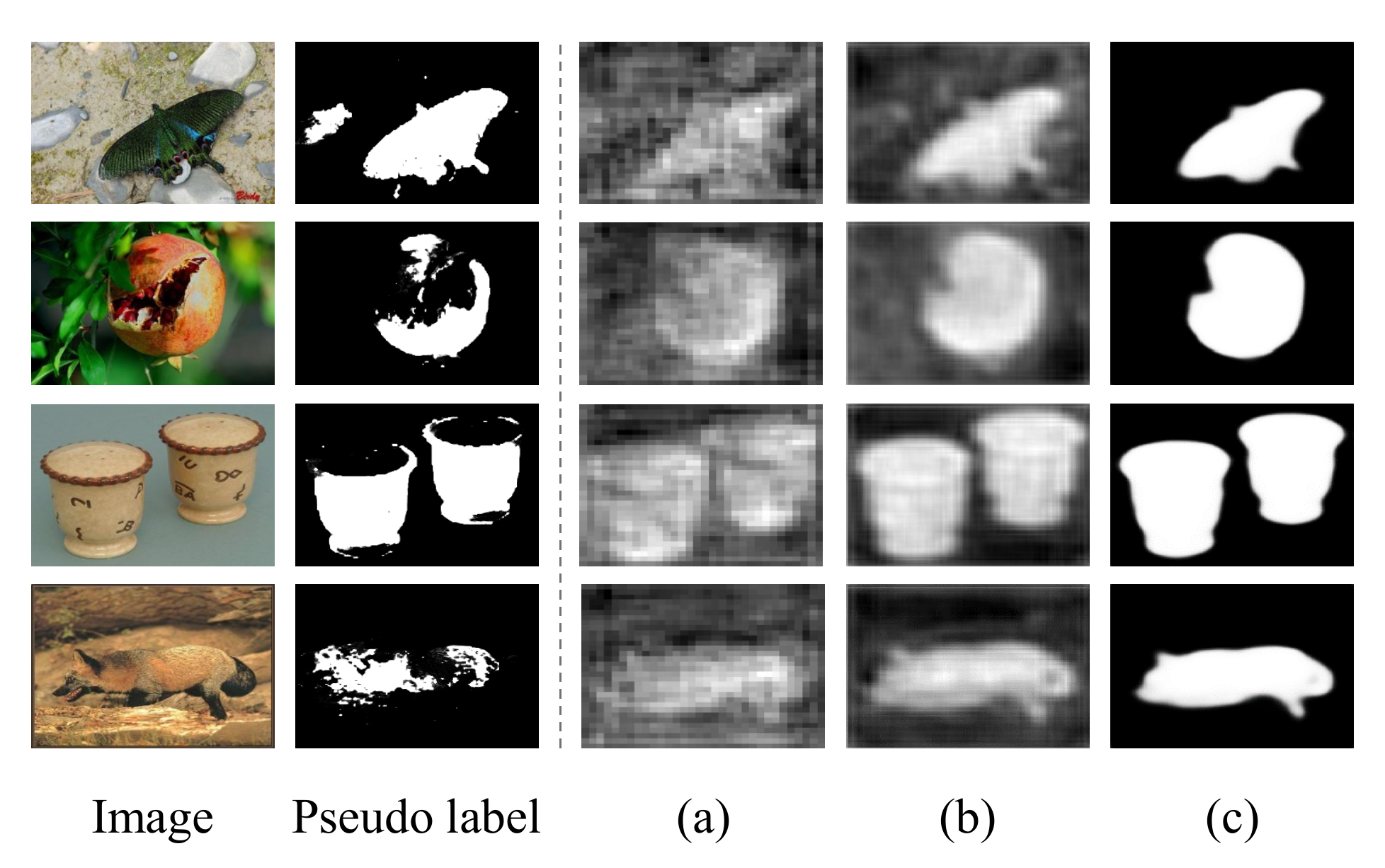}
\vspace{-4mm}
\caption{ Visualization of the directive filter $F1$. (a) and (b) represent the feature maps from the $2^{nd}$ and $4^{th}$ convolution layers of the directive filter, and (c) indicates the predictions $P_1$ of $F1$.}
\label{feature}
\vspace{-3mm}
\end{figure}

To be specific, we first design a directive filter (DF) to extract and filter the more accurate saliency cues from pseudo labels. The architecture of the proposed directive filters is illustrated in Figure {\color{red}\ref{framework}}. It takes the features from the shared encoder as input, and extracts the saliency cues from pseudo labels through several convolution  layers. 
As is pointed out in~\cite{Fan2020EmployingMF, Araslanov2020SingleStageSS, Veksler2020RegularizedLF, Fan2020LearningIO}, the convolutional neural networks possess good robustness to noisy labels. Therefore, the inaccurate saliency cues in pseudo labels can be gradually corrected by the convolution layers in DF.
As shown in Figure {\color{red}\ref{feature}}, the extra noise and incomplete defects in pseudo labels are progressively corrected, and more concrete saliency cues are extracted through convolutions.

To effectively utilize and integrate the comprehensive saliency cues from multiple pseudo labels, we design the proposed MFNet as is illustrated in Figure {\color{red}\ref{framework}}. Firstly, we introduce two directive filters $F1$ and $F2$ to filter and extract accurate saliency cues from pseudo labels $Y_1$ and $Y_2$ respectively. To attach equal importance to different pseudo labels, we set the same settings for two directive filters. The corresponding training loss $L_1$ and $L_2$ for $F1$ and $F2$ are computed as:

\vspace{-5mm}
\begin{equation}
\begin{split}
{L_k}({P_k},{Y_k}) = & - \sum\limits_{\rm{i}} {{\; y_{ki}}} *\log {p_k}_i - (1 - {y_{ki}})* \\
& \log (1 - {p_{ki}}),\quad k = 1,2,
\end{split}
\end{equation}
\vspace{-3mm}

\noindent where $p_{ki}$ and $y_{ki}$ represent the elements of the directive filter predictions $P_k$ and its pseudo labels $Y_k$.
Secondly, we then simultaneously propagate these filtered accurate cues to the saliency decoder through a multi-guidance loss $L_{mg}$, which can be described as follows:

\vspace{-3mm}
\begin{equation}
\begin{split}
{L_{\rm{mg}}}({P_s},{Y_s}) = & - \sum\limits_i {(1 - {y_i})*\log (1 - {p_{si}})} 
\\&- {y_i}*\log {p_{si}} ,
\end{split}
\end{equation}
\vspace{-3mm}

\noindent where $p_{si}$ is the elements of the saliency decoder prediction $P_s$. $Y_s$ is the average prediction of directive filters after the aforementioned pixel-wise refinement~\cite{Araslanov2020SingleStageSS}, and $y_i$ is its elements.

In addition, we adopt a self-supervision strategy between two directive filters, which aims to encourage two filters to extract similar saliency cues from different pseudo labels. The insight is that the common saliency cues learned from different pseudo labels describe more accurate and authentic saliency information. The loss $L_{ss}$ of this self-supervision term is defined as follows:

\vspace{-2mm}
\begin{equation}
\begin{split}
{L_{ss}}({P_1},{P_2}) =  - {\sum\limits_i {({p_{1i}} - {p_{2i}})} ^2}.
\end{split}
\end{equation}
\vspace{-4mm}

The final loss function $L$ for training the proposed MFNet is given by the combination of the above loss functions:

\vspace{-6mm}
\begin{equation}
\begin{split}
L = {L_1} + {L_2} + {L_{\rm{mg}}} + \delta {L_{ss}},
\end{split}
\end{equation}
\vspace{-4mm}

\noindent where $\delta$ is a hyper-parameter which controls the weight of the self-supervision term.

\begin{table*}[!t]
	\centering
	\small
	\renewcommand\arraystretch{1.3}  
	\setlength{\tabcolsep}{0.25mm}
	\vspace{-1mm}
	\begin{threeparttable}
	\caption{ Quantitative comparisons of E-measure ($E_s$), S-measure ($S_{\alpha}$), F-measure ($F_{\beta}$) and MAE ($M$) metrics over five benchmark datasets. The supervision type (\textbf{Sup.}) I indicates using category annotations only, and I\&C represents developing WSOD on both category and caption annotations simultaneously. - means unavailable results. The best results are marked in \textbf{boldface}.}

	\label{quantitative}
    \begin{tabular}{ccp{0.7cm}<{\centering}p{0.7cm}<{\centering}p{0.7cm}<{\centering}p{0.7cm}<{\centering}p{0.7cm}<{\centering}p{0.7cm}<{\centering}p{0.7cm}<{\centering}p{0.7cm}<{\centering}p{0.7cm}<{\centering}p{0.7cm}<{\centering}p{0.7cm}<{\centering}p{0.7cm}<{\centering}p{0.7cm}<{\centering}p{0.7cm}<{\centering}p{0.7cm}<{\centering}p{0.7cm}<{\centering}p{0.7cm}<{\centering}p{0.7cm}<{\centering}p{0.7cm}<{\centering}p{0.7cm}<{\centering}}
    \toprule
    \multicolumn{1}{c}{\multirow{2.5}{*}{Methods  }}&
    \multicolumn{1}{c}{\multirow{2.5}{*}{ Sup.}}&
    \multicolumn{4}{c}{ECSSD}& \multicolumn{4}{c}{DUTS-Test}& \multicolumn{4}{c}{HKU-IS}& \multicolumn{4}{c}{DUT-OMRON}& \multicolumn{4}{c}{PASCAL-S}\cr
    \cmidrule(lr){3-6} \cmidrule(lr){7-10}\cmidrule(lr){11-14}\cmidrule(lr){15-18}\cmidrule(lr){19-22}
    &{}&$S_{\alpha}$		&$E_s$		 &$F_{\beta}$ 	&$M$  	&$S_{\alpha}$		&$E_s$	 &$F_{\beta}$ 	&$M$  &$S_{\alpha}$	&$E_s$			 &$F_{\beta}$ 	&$M$ &$S_{\alpha}$ 	&$E_s$			 &$F_{\beta}$ 	&$M$  &$S_{\alpha}$		&$E_s$		 &$F_{\beta}$ 	&$M$\cr

	\midrule
	\multicolumn{1}{c}{\multirow{1}{*}{WSS~\cite{Wang2017LearningTD}}} &I	 
	&.811 		&.869 		&.823 		&.104 			
	&.748 		&.795 		&.654 		&.100  		
	&.822 		&.896 		&.821 		&.079 		
	&.725 		&.768 		&.603 		&.109 		
	&.744 		&.791 		&.715 		&.139		\cr 
 
	\multicolumn{1}{c}{\multirow{1}{*}{ASMO~\cite{Li2018WeaklySS}}} &I
	&.802 		&.853 		&.797 		&.110 			
	&.697 		&.772 		&.614 		&.116 		
	&- 			&- 			&- 			&- 			
	&.752 		&.776 		&.622 		&.101 		
	&.717 		&.772 		&.693 		&.149		\cr

	\multicolumn{1}{c}{\multirow{1}{*}{MSW}~\cite{Zeng2019MultiSourceWS}} & I\&C
	&.827 		&.884 		&.840 		&.096		
	&.759 		&.814 		&.684 		&.091 		
	&.818 		&.895 		&.814 		&.084 		
	&\bf{.756} 	&.763 		&.609 		&.109 		
	&.768 		&.790 		&.713 		&.133		\cr

	\midrule   
	\multicolumn{1}{c}{\multirow{1}{*}{\textbf{MFNet}}} &I 
	&\bf{.834} 	&\bf{.885} 	&\bf{.854} 	&\bf{.084}		
	&\bf{.775} 	&\bf{.839} 	&\bf{.710}	&\bf{.076} 	
	&\bf{.846} 	&\bf{.921} 	&\bf{.851} 	&\bf{.059} 	
	&.742 			&\bf{.803} 	&\bf{.646} 	&\bf{.087} 	
	&\bf{.770} 	&\bf{.817} 	&\bf{.751} 	&\bf{.115}	\cr

	\bottomrule
    \end{tabular}
    \end{threeparttable}
    \vspace{-2mm}
\end{table*}

\begin{figure*}[!t]
\vspace{-2mm}
\includegraphics[width=1.00\linewidth]{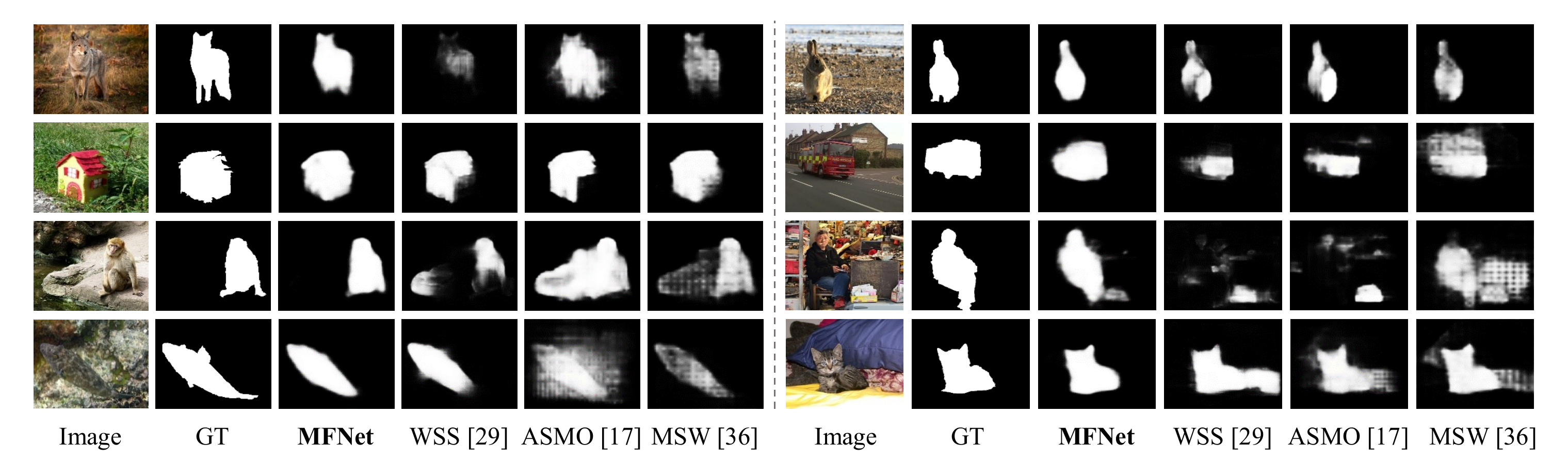}
\vspace{-4mm}
\caption{ Visual comparisons of our method with existing image-level annotation based WSOD methods in some challengling scenes.}
\label{qualitative}
\vspace{-4mm}
\end{figure*}

The architecture of the saliency network is illustrated in Figure {\color{red}\ref{framework}}. We adopt a simple encoder-decoder framework, which usually served as baseline network in SOD. It takes three features from the $3^{rd}$, $4^{th}$ and $5^{th}$ convolution blocks of the encoder as input, and perform multi-scale bottom-up aggregation~\cite{Zhang2017AmuletAM}. The predictions $P_s$ of the saliency decoder is our final prediction. During testing, we only retain the saliency network and discard the multiple directive filters for acceleration.
\section{Experiments}

\subsection{Implementation Details}
We conduct our method on the Pytorch toolbox with a RTX 2080Ti GPU. The shared encoder in our method is designed based on DenseNet-169~\cite{Huang2017DenselyCC}, which is same as the latest work MSW~\cite{Zeng2019MultiSourceWS}. During the training phase of the classification network, we adopt the Adam optimization algorithm~\cite{2014Adam} and set the learning rate and the maximum iteration to 1e-4 and 20000, respectively. In the inference phase, we generate CAMs using a multi-inference strategy following the settings of~\cite{Ahn2018LearningPS}. To be specific, the input images are flipped and then resized to four scales. The final maps are computed as the average of corresponding eight CAMs. For the saliency network, we only take the RGB images of DUTS-Train dataset~\cite{Wang2017LearningTD} and the generated pseudo labels for training. In this stage, we also adopt the Adam optimization algorithm and set the learning rate and the maximum iteration to 3e-6 and 26000, respectively. All the training images are resized to $256\times 256$ and the parameters of new-added layers are initialized by Xavier algorithm~\cite{pmlr-v9-glorot10a}. The source code will be released upon publication.

\subsection{Datasets and Evaluation Metrics}
Following the previous works~\cite{Wang2017LearningTD, Zeng2019MultiSourceWS}, we adopt ImageNet~\cite{5206848} and DUTS-Train dataset~\cite{Wang2017LearningTD} as our training sets for the classification network and the proposed MFNet respectively for the sake of fairness. We test our method on five widely-adopted datasets: ECSSD~\cite{Yan2013HierarchicalSD}, DUTS-Test~\cite{Wang2017LearningTD}, HKU-IS~\cite{Li2015VisualSB}, DUT-OMRON~\cite{2013Saliency} and PASCAL-S~\cite{Li2014TheSO}. \textbf{ECSSD} contains 1000 images of different sizes with obvious salient objects. \textbf{DUTS-Test} includes 5019 samples of various challenging scenes. \textbf{HKU-IS} consists of 4447 images with many multiple-object scenes. \textbf{DUT-OMRON} contains 5168 images with complex structures and contours. \textbf{PASCAL-S} includes 850 samples that are annotated by 8 subjects in eye-tracking tests.

\begin{figure*}
\vspace{-3mm}
\includegraphics[width=1.00\linewidth]{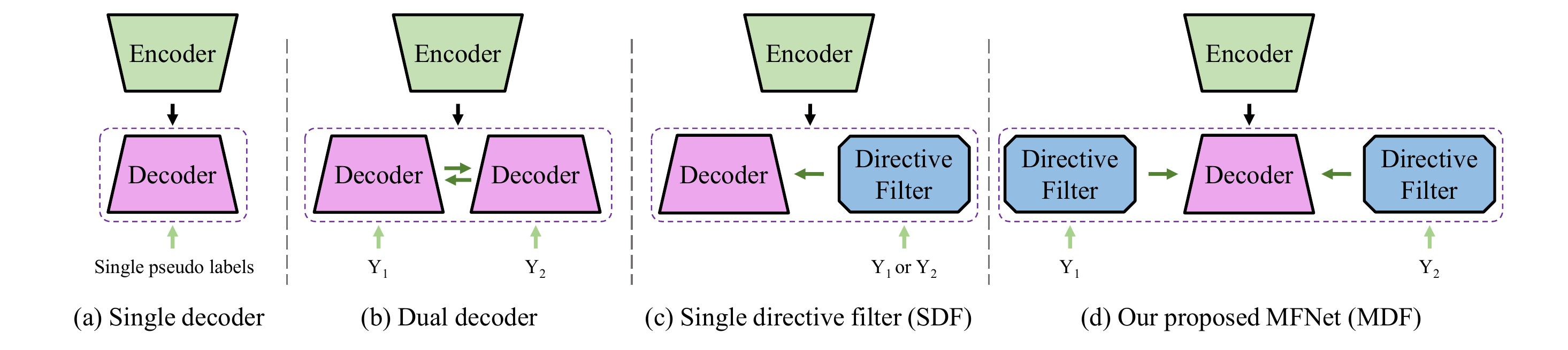}
\vspace{-4mm}
\caption{ The frameworks of different settings in ablation studies. (a) indicates single pseudo label cases (1) to (2) and (5) to (7), (b) refers to dual-decoder framework in case (8), and (c) indicates single directive filter (SDF) cases (3) and (4). (d) is our proposed MFNet using multiple directive filters (MDF), which corresponds to case (9). }
\label{ablation_set}
\vspace{-1mm}
\end{figure*}

For a comprehensive comparison, we adopt four well-accepted metrics, including S-measure~\cite{Fan2017StructureMeasureAN}, E-measure~\cite{ijcai2018-97}, F-measure~\cite{5206596} as well as Mean Absolute Error (MAE), to evaluate our method. Specifically, S-measure focuses on evaluating the structural information of saliency maps and evaluates region-aware and object-aware structural similarity between saliency maps and ground truths. E-measure attaches more importance on the unification of global and local information. Besides, F-measure is a harmonic mean of average precision and average recall, and MAE evaluates the average difference between saliency maps and ground truths.

\subsection{Comparison with State-of-the-arts}

We compare our approach denoted as \textbf{MFNet} with the existing image-level category label based WSOD methods: WSS~\cite{Wang2017LearningTD}, ASMO~\cite{Li2018WeaklySS} and MSW~\cite{Zeng2019MultiSourceWS}. The quantitative and qualitative results are illustrated in the Table {\color{red}\ref{quantitative}} and Figure {\color{red}\ref{qualitative}}. For a fair comparison, we obtain the saliency maps of these methods from authors and conduct same evaluation on all the methods.

\begin{spacing}{1.3}
\end{spacing}
\noindent \textbf{Quantitative evaluation.} 
The quantitative results on five datasets are shown in Table {\color{red}\ref{quantitative}}. It can be seen that our method outperforms all the previous works on almost metrics except for S-measure on the DUT-OMRON dataset. It is worth noting that F-measure of our method is significantly better than the second best results on PASCAL-S (0.751 against 0.713), HKU-IS (0.851 against 0.814) and DUT-Test (0.710 against 0.684). The improvements on MAE metrics further prove the superiority of our method. Especially, 29.7\% improvement on HKU-IS dataset and 20.2\% on DUT-OMRON dataset are achieved. 
\textbf{Moreover}, from a deeper perspective, the previous work ASMO~\cite{Li2018WeaklySS} achieves better performance on the challenging DUT-OMRON dataset while WSS~\cite{Wang2017LearningTD} and MSW~\cite{Zeng2019MultiSourceWS} show more superiority on the other datasets. This is because the former uses a traditional SOD method MB+~\cite{Zhang2015MinimumBS} to perform refinement and generate pseudo labels, while the latter leverages the aforementioned superpixel-wise refinement. It demonstrates that the prejudiced single pseudo label from different refinement algorithms does lead to different generalization abilities of WSOD methods. Based on these observations, we argue that exploring multiple pseudo labels is necessary and the results in Table {\color{red}\ref{quantitative}} also prove its effectiveness.

\begin{spacing}{1.3}
\end{spacing}
\noindent \textbf{Qualitative evaluation.} 
Figure {\color{red}\ref{qualitative}} shows the qualitative comparisons of our MFNet with existing WSOD methods in some challenging scenes. It can be seen that our method could segment more accurate and integrated objects than other methods. For example, in some similar foreground and background scenes in the $1^{st}$, $3^{rd}$ and $4^{th}$ rows on the left in Figure {\color{red}\ref{qualitative}}, our method could discriminate more salient objects accurately from its similar background. When the background comes complex and noisy such as in the $2^{nd}$ and $3^{rd}$ rows on the right, our method could also perform better than the others.

\begin{table*}[!t]
	\renewcommand\arraystretch{1.1}  
	\small
  	\centering
  	\setlength{\tabcolsep}{0.95mm}
	\vspace{0mm}
  	\begin{threeparttable}
	\caption{ Quantitative results of ablation studies, \textbf{Type} means the number of used pseudo labels and \textbf{Pseudo label} indicates different pseudo labels Y$_1$ and Y$_2$. \textbf{DF} represents our proposed directive filter (DF). In \textbf{Case}: (1) and (2) indicate the case which trains the saliency networks with $Y_1$ and $Y_2$ respectively. Based on (1) and (2), case (3) and (4) adopt the proposed DF. Cases (5) to (7) first integrate multiple labels through average (Avg($\cdot$)), intersection ($\cap$) and union ($\cup$) respectively, and then train the saliency networks on these integrated labels. Case (8) adopts a straightforward dual-decoder framework and case (9) is our final MFNet.}
	\label{ablation}
	\begin{tabular}{ccp{1.0cm}<{\centering}p{1.7cm}<{\centering}p{1.0cm}<{\centering}p{1.0cm}<{\centering}p{1.0cm}<{\centering}p{1.0cm}<{\centering}p{1.0cm}<{\centering}p{1.0cm}<{\centering}p{1.0cm}<{\centering}p{1.0cm}<{\centering}p{1.0cm}<{\centering}p{1.0cm}<{\centering}p{1.0cm}<{\centering}p{1.0cm}<{\centering}}

	\toprule
	\multicolumn{1}{c}{\multirow{2}{*}{\normalsize Type   }}&
	\multicolumn{1}{c}{\multirow{2}{*}{\normalsize Case     }}&
	\multicolumn{1}{c}{\multirow{2}{*}{\normalsize \,DF }}&
	\multicolumn{1}{c}{\multirow{1.5}{*}{\normalsize Pseudo}}&
	\multicolumn{2}{c}{ECSSD}& 
	\multicolumn{2}{c}{DUTS-Test}& 
	\multicolumn{2}{c}{HKU-IS} & 
	\multicolumn{2}{c}{DUT-OMRON}& 
	\multicolumn{2}{c}{PASCAL-S}\cr
	\cmidrule(lr){5-6} \cmidrule(lr){7-8} \cmidrule(lr){9-10} \cmidrule(lr){11-12} \cmidrule(lr){13-14} 
	&{ }	&{   }	 &{\normalsize label}	  
	&$F_{\beta}$$\uparrow$ 		&$M$$\downarrow$
	&$F_{\beta}$$\uparrow$ 		&$M$$\downarrow$
	&$F_{\beta}$$\uparrow$ 		&$M$$\downarrow$
	&$F_{\beta}$$\uparrow$ 		&$M$$\downarrow$
	&$F_{\beta}$$\uparrow$ 		&$M$$\downarrow$	\cr
	\midrule
	\multicolumn{1}{c}{\multirow{4}{*}{Single}}
   	
	& (1)	& 					& Y$_1$		
	& 0.818 		& 0.113 		& 0.607		& 0.099		& 0.824		& 0.080		& 0.607  		& 0.099		& 0.724		& 0.134		\cr

	& (2)	& 			 		& Y$_2$		
	& 0.824		& 0.090		& 0.639		& 0.090		& 0.801		& 0.067		& 0.576		& 0.108		& 0.717		& 0.122		\cr

	& (3)	&  \checkmark	& Y$_1$		
	& 0.835		& 0.095		& 0.698		& 0.082		& 0.840		& 0.066		& 0.641		& 0.089		& 0.734		& 0.125		\cr

	& (4)	& \checkmark		& Y$_2$		
	& 0.847		& 0.085		& 0.684		& 0.084		& 0.836		& 0.062		& 0.602		& 0.103		& 0.743		& \bf0.115	\cr

	\midrule	
	\multicolumn{1}{c}{\multirow{5}{*}{Multiple}}
	& (5)	&  				&  \small{Avg(Y$_1$, Y$_2$)}
	& 0.826		& 0.087		& 0.638		& 0.088		& 0.800		& 0.066		& 0.576		& 0.106		& 0.716		& 0.120		\cr

	& (6)	&  				& Y$_1$ $\cap$ Y$_2$			
	& 0.831		& 0.086		& 0.649		& 0.085		& 0.810		& 0.064		& 0.595		& 0.098		& 0.723		& 0.118		\cr

	& (7)	&  				& Y$_1$ $\cup$ Y$_2$				
	& 0.823		& 0.091		& 0.637		& 0.093		& 0.800		& 0.070		& 0.637		& 0.093		& 0.714		& 0.124		\cr

	& (8)	&  				& Y$_1$ \& Y$_2$			
	& 0.843		& 0.087		& 0.670		& 0.083		& 0.831		& 0.064		& 0.607		& 0.093		& 0.735		& 0.118		\cr

	& (9)	& \checkmark	& Y$_1$ \& Y$_2$		
	& \bf0.854		& \bf0.084		& \bf0.710		& \bf0.076		& \bf0.851		
	& \bf0.059		& \bf0.646		& \bf0.087		& \bf0.751		& \bf0.115	\cr
	\bottomrule
	\end{tabular}
	\end{threeparttable}
	\vspace{-3mm}
\end{table*}
\subsection{Ablation Studies}

We design various cases in ablation studies to prove the superiority of our method comprehensively. For a clearer description, the different frameworks of each case in Table {\color{red}\ref{ablation}} are shown in Figure {\color{red}\ref{ablation_set}}.

\begin{spacing}{1.3}
\end{spacing}
\noindent \textbf{Effectiveness of Directive Filter. } 
We propose a directive filter (DF) to extract and filter more accurate saliency cues from noisy pseudo labels. It can be applied to both single pseudo label setting (SDF) and multiple setting (MDF) according to the numbers of pseudo labels. 
\textbf{On the one hand}, SDF can encourage promising improvements on all datasets as shown in cases (1) to (4) in Table {\color{red}\ref{ablation}}, especially on two challenging datasets DUTS-Test and DUT-OMRON.
It indicates that when pseudo labels tend to be more inaccurate and noisy in challenging scenes, normal saliency networks inevitably learn more negative information from its direct supervision. In these scenes, the proposed SDF can filter and extract accurate saliency cues and then encourages a more powerful saliency decoder. 
\textbf{On the other hand}, MDF can effectively integrate multiple saliency cues in various pseudo labels.
To prove its superiority, we design four different cases to fuse multiple saliency cues, including three simple ways: average (Avg($\cdot$)), intersection ($\cap$) and union ($\cup$), as well as the aforementioned straightforward way: dual decoder. The results in cases (5) to (7) prove that such three simple ways cannot adequately enough to leverage multiple information. The better performance of case (8) indicates that a more proper approach to leverage multiple labels can achieve a promising improvement. Cases (9) is our final MFNet with MDF shown in (d) in Figure {\color{red}\ref{ablation_set}}, it can be seen that MDF contributes to ourperforming all the other multiple settings by a large margin, especially on two challenging datasets DUTS-Test and DUT-OMRON. These observations support: 1) the effectiveness of our proposed DF on extracting accurate saliency cues. 2) the superiority of the proposed MDF on integrating multiple saliency cues.
\textbf{Moreover}, as is illustrated in Table {\color{red}\ref{guidance_main}}, the saliency decoder achieves an obvious improvement compared to its directive filters (DFs). It proves that the filtered saliency cues from DFs are accurate enough to encourage better results with the proposed multi-guidance loss.

\begin{table}[!t]
	\renewcommand\arraystretch{1.2}  
	\small
  	\centering
  	\setlength{\tabcolsep}{1.02mm}
	\vspace{1mm}
  	\begin{threeparttable}
	\caption{ Comparisons on the results of the saliency decoder and its two directive filters. Supervised by more accurate saliency cues from directive filters, the final saliency decoder achieves promising improvements.}
	\label{guidance_main}
	\begin{tabular}{ccp{1cm}<{\centering}p{0.9cm}<{\centering}p{0.9cm}<{\centering}p{0.9cm}<{\centering}p{0.9cm}<{\centering}p{0.9cm}<{\centering}p{0.9cm}<{\centering}p{0.9cm}<{\centering}}
	\toprule
    \multicolumn{1}{c}{  }&
    \multicolumn{1}{c}{\multirow{2}{*}{Results}}&
    \multicolumn{2}{c}{ECSSD}&
    \multicolumn{2}{c}{DUTS-Test}&
    \multicolumn{2}{c}{HKU-IS} \cr
    \cmidrule(lr){3-4} \cmidrule(lr){5-6} \cmidrule(lr){7-8} 
    & {} &$F_{\beta}$$\uparrow$ 	&$M$$\downarrow$  &$F_{\beta}$$\uparrow$ 	&$M$$\downarrow$ 	&$F_{\beta}$$\uparrow$ 	&$M$$\downarrow$ 	\cr

	\midrule
	& ${P_1}$		& 0.842		& 0.089		& 0.689		& 0.079		& 0.836		& 0.063		\cr
   
	& ${P_2}$		& 0.844		& 0.088		& 0.686 		& 0.080 		& 0.836		& 0.063		\cr

	& Final ${P_s}$		& \bf0.854	& \bf0.084	& \bf0.710	& \bf0.076	& \bf0.851	& \bf0.059	\cr

	\bottomrule
	\end{tabular}
	\end{threeparttable}
	\vspace{-5mm}
\end{table}

\begin{figure}
\vspace{1mm}
\includegraphics[width=1.00\linewidth]{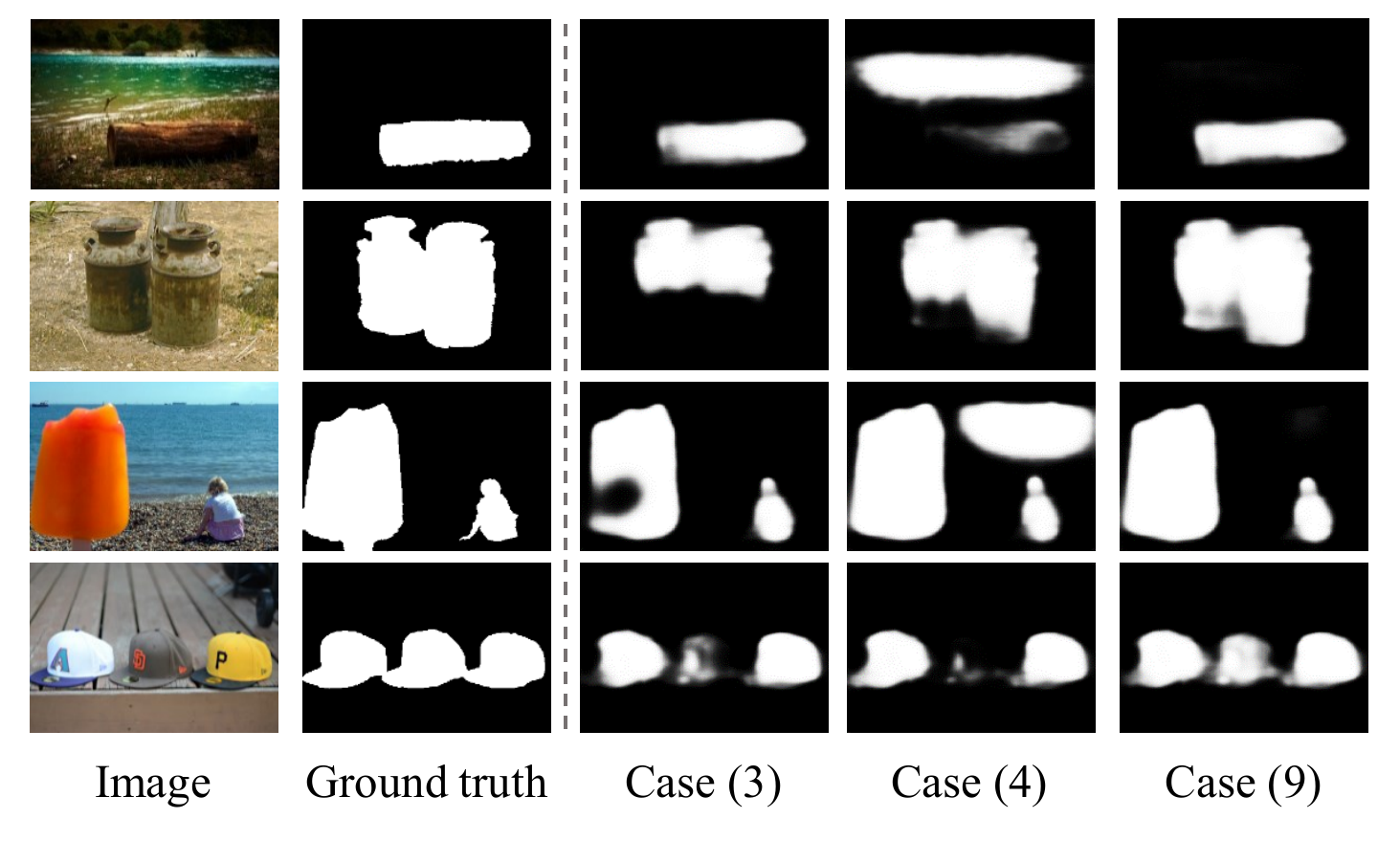}
\setlength{\abovecaptionskip}{0.2cm}
\setlength{\belowcaptionskip}{0.2cm}
\vspace{-4mm}
\caption{ Visual analysis of the effectiveness of multiple pseudo labels. Case(3), (4) and (9) represent the results of cases (3), (4) and (9), respectively. It can be seen that multiple labels encourage more generalized and accurate results compared to single label. }
\label{ablation_visual}
\vspace{-5mm}
\end{figure}

\begin{spacing}{1.3}
\end{spacing}
\noindent \textbf{Effectiveness of multiple pseudo labels. } 
We introduce a multiple-pseudo-label WSOD framework, which targets to integrate multiple saliency cues to avoid the bias of each single pseudo label.
\textbf{First of all}, as is mentioned above, cases (5) to (7) in Table {\color{red}\ref{ablation}} are the aforementioned simple ways to integrate multiple cues. Cases (5) and (7) lead to similar performance and do not get obvious improvements than single pseudo label cases (1) and (2), while case (6) achieves good improvements especially on the MAE metric. These results indicate that the average and union introduce more redundant noises from both pseudo labels and lead to inferior performance. The reason why case (6) achieves better performance than cases (5) and (7) is that the intersection operation on two pseudo labels can help to generates high-confidence labels. 
\textbf{Moreover}, by adopting the dual-decoder framework in case (8), a remarkable improvement is achieved over the single pseudo label case (1) and (2), which proves the superiority of multiple pseudo labels and inspires us for a further exploration.
\textbf{Last but not least}, case (9) is our proposed MFNet, compared to cases (3) and (4), a promising improvement is achieved on all metrics, which furthur proves the superiority of multiple cues.
Figure {\color{red}\ref{ablation_visual}} provides the visual results of multiple DF and single DF settings. It proves that the more comprehensive saliency cues in multiple pseudo labels helps to avoid the negative impacts from single label and encourage more robust results.


\begin{table}[!t]
	\renewcommand\arraystretch{1.2}  
	\small
  	\centering
  	\setlength{\tabcolsep}{0.82mm}
	\vspace{-1mm}
  	\begin{threeparttable}
	\caption{ The experiments on the effect of self-supervision and the setting of its hyper-parameter $\delta$. The best and second-best results are marked in \textbf{boldface} and \underline{underline}, respectively.}
	\label{self-supervision}
	\begin{tabular}{cp{1.0cm}<{\centering}p{1cm}<{\centering}p{1cm}<{\centering}p{1cm}<{\centering}p{1cm}<{\centering}p{1cm}<{\centering}p{1cm}<{\centering}p{1cm}<{\centering}}
	\toprule
    \multicolumn{1}{c}{  }&
    \multicolumn{1}{c}{\multirow{2}{*}{\normalsize$\delta$}     }&
    \multicolumn{2}{c}{ ECSSD}&
    \multicolumn{2}{c}{ DUTS-Test}&
    \multicolumn{2}{c}{ HKU-IS} \cr
    \cmidrule(lr){3-4} \cmidrule(lr){5-6} \cmidrule(lr){7-8} 
    & { } &$F_{\beta}$$\uparrow$	&$M$$\downarrow$  &$F_{\beta}$$\uparrow$ 	&$M$$\downarrow$ 	&$F_{\beta}$$\uparrow$ 	&$M$$\downarrow$ 	\cr

	\midrule
	& {-2   }	
	& {0.844}					& \bf0.081				& 0.679					& 0.083					& 0.837					& \bf0.058				\cr
   
	& {0    }		
	& \underline{0.851}		& \underline{0.084}		& 0.702					& \underline{0.077}		& 0.848					& \bf0.058				\cr

	& {$\rightarrow$2$\leftarrow$}		
	& \bf0.854				& \underline{0.084}		& \bf0.710				& \bf0.076				& \bf0.851				& \underline{0.059}		\cr

	& {4    }		
	& {0.848}					& 0.089					& \underline{0.706}		& 0.078					& \underline{0.850}		& 0.061					\cr

	\bottomrule
	\end{tabular}
	\end{threeparttable}
	\vspace{-3mm}
\end{table}

\subsection{Hyper-parameter Settings}
\begin{spacing}{1.2}
\end{spacing}
We adopt a self-supervision strategy between multiple directive filters, aiming to force them to learn more authentic saliency cues from various pseudo labels. For a comprehensive comparison, we set the hyper-parameter $\delta$ from -2 to 4 in Table {\color{red}\ref{self-supervision}} to discuss the effectiveness of the self-supervision strategy as well as the hyper-parameter $\delta$. 

To be specific, when the $\delta$ is set to -2, the directive filters are encouraged to learn different saliency cues from various pseudo labels. Setting $\delta$ to 0 means that we do not adopt the self-supervision strategy, and the last two rows in Table {\color{red}\ref{self-supervision}} indicate different hyper-parameters for the self-supervision strategy. It can be seen that encouraging multiple directive filters to learn similar cues does perform better than the other settings and the best performance is achieved when $\delta$ is set to 2.

\subsection{Application}
To further demonstrate the effectiveness of our proposed framework, we extend it to the latest WSOD methods MSW~\cite{Zeng2019MultiSourceWS}. To be specific, for the coarse maps generated from the multi-source weak supervisions, we also perform two different refinements as we do to synthesize different pseudo labels, and then adopt the proposed multiple-pseudo-label framework to extract and integrate the multiple saliency cues. The architecture of the saliency decoders follows the original setting in MSW for the fair comparison.
Here, we add weighted F-measure $F_{\beta}^{\omega}$~\cite{Margolin2014HowTE} for a more comprehensive comparison. 

\begin{figure}[!t]
\vspace{-1mm}
\includegraphics[width=1.00\linewidth]{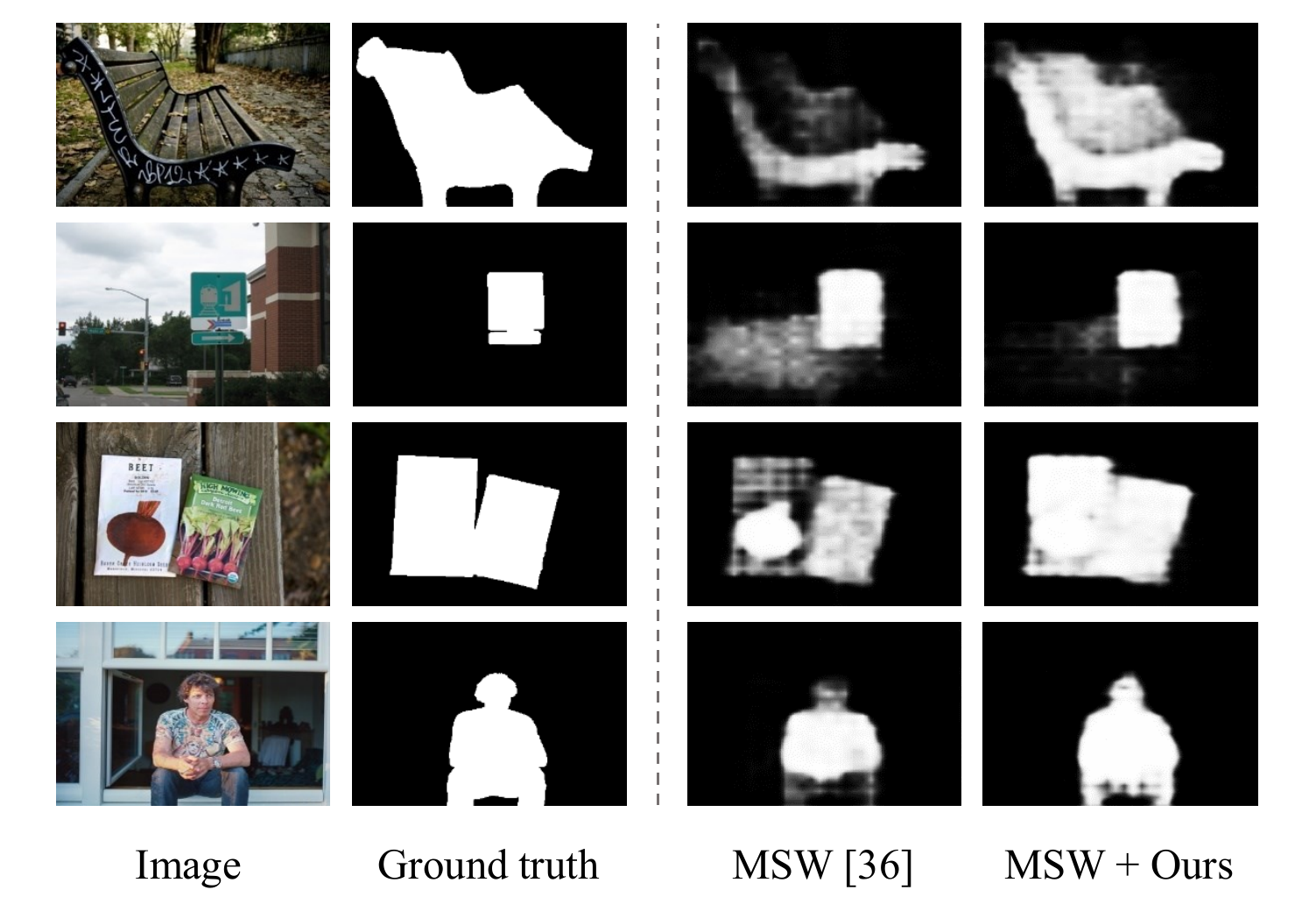}
\vspace{-4mm}
\caption{ Visual analysis of applying our framework on the latest previous work MSW~\cite{Zeng2019MultiSourceWS}.}
\label{application_fig}
\vspace{-4mm}
\end{figure}

The results in Table {\color{red}\ref{application_tab}} illustrate that remarkable improvements are achieved especially on the $F_{\beta}^{\omega}$ and MAE metrics. It indicates that the proposed multiple pseudo label framework does adequately integrate saliency cues from multiple labels and help existing method to achieve better performance. The visual analysis in Figure {\color{red}\ref{application_fig}} also supports this observation, in which our framework helps MSW to predict more accurate and complete saliency maps even in challenging scenes.
\textbf{Moreover}, it is worth noting that the proposed framework can not only be extended to other single pseudo label methods, but also flexible enough to integrate more other pseudo labels by just adding more directive filters when more pseudo labels can be obtained.

\begin{table}[!t]
	\renewcommand\arraystretch{1.2}  
	\small
  	\centering
  	\setlength{\tabcolsep}{0.45mm}
	\vspace{-1mm}
  	\begin{threeparttable}
	\caption{ The experiments of applying our multiple-pseudo-label framework on the latest work MSW~\cite{Zeng2019MultiSourceWS}. }
	\label{application_tab}
	\begin{tabular}{ccp{1cm}<{\centering}p{1cm}<{\centering}p{1cm}<{\centering}p{1cm}<{\centering}p{1cm}<{\centering}p{1cm}<{\centering}p{1cm}<{\centering}}
	\toprule
    \multicolumn{1}{c}{  }&
    \multicolumn{1}{c}{\multirow{2}{*}{Settings  }}&
    \multicolumn{3}{c}{ ECSSD}&
    \multicolumn{3}{c}{ HKU-IS} \cr
    \cmidrule(lr){3-5} \cmidrule(lr){6-8} 
	& { }  &$F_{\beta}$$\uparrow$   &$F_{\beta}^{\omega}$$\uparrow$	 &$M$$\downarrow$  
	&$F_{\beta}$$\uparrow$   &$F_{\beta}^{\omega}$$\uparrow$	 &$M$$\downarrow$ 	\cr

	\midrule
	& {MSW~\cite{Zeng2019MultiSourceWS}  }			
	& 0.840		& 0.716		& 0.096			& 0.814		& 0.685		& 0.084			\cr
	& {$+$ Ours  }	
	& \bf+0.016	& \bf+0.065	& \bf-0.019		& \bf+0.006	& \bf+0.058	& \bf-0.015	\cr
	\bottomrule
	\end{tabular}
	\end{threeparttable}
	\vspace{-1mm}
\end{table}

\section{Conclusion}
In this paper, we propose to utilize multiple pseudo labels to avoid the negative impacts from the prejudiced single label. To this end, we introduce a new framework to explore more comprehensive and accurate saliency cues from multiple labels. To be specific, we design a multi-filter directive network (MFNet) which consists of an encoder-decoder saliency network as well as multiple directive filters. We first use multiple directive filters to extract and filter more accurate saliency cues from multiple labels, and then propagate these filtered cues to the saliency decoder simultaneously. We also adopt a self-supervision strategy to encourage similar guidance of different directive filters, and implicitly integrate multiple saliency cues with a multi-guidance loss. Comparisons with previous methods prove the superiority of the proposed method, and ablation studies also support the effectiveness of each component.

\begin{spacing}{2.0}
\end{spacing}
\noindent \textbf{Acknowledgements.} This work was supported by the Science and Technology Innovation Foundation of Dalian (\#2019J12GX034), the National Natural Science Foundation of China (\#61976035), and the Fundamental Research Funds for the Central Universities (\#DUT20JC42).

{\small
\bibliographystyle{ieee_fullname}
\bibliography{egbib}
}

\end{document}